\documentclass[final]{cvpr}

\usepackage{times}
\usepackage{epsfig}
\usepackage{graphicx}
\usepackage{amsmath}
\usepackage{amssymb}

\usepackage{color}
\usepackage{booktabs}
\usepackage{multirow}
\usepackage{rotating}
\usepackage{pifont}
\usepackage{fixmath}
\usepackage{siunitx}
\usepackage{etoolbox}
\usepackage[caption=false]{subfig}
\usepackage{algpseudocode}
\usepackage[numbers,sort]{natbib}

\usepackage[pagebackref=true,breaklinks=true,colorlinks]{hyperref}

\MakeRobust{\Call}
\robustify\bfseries

\renewcommand{\figref}[1]{Figure~\ref{#1}}
\renewcommand{\tabref}[1]{Table~\ref{#1}}
\renewcommand{\eqref}[1]{Eq.~(\ref{#1})}

\newcommand{\cmark}{\ding{51}}
\newcommand{\xmark}{\ding{55}}

\newcommand{\PAR}[1]{\smallskip \noindent \textbf{#1}}
\newcommand{\PARbegin}[1]{\noindent \textbf{#1}}

\begin{document}

\title{Learning to Recommend Frame for\\Interactive Video Object Segmentation in the Wild}

\author{
Zhaoyuan Yin\textsuperscript{\rm 1},
Jia Zheng\textsuperscript{\rm 2},
Weixin Luo\textsuperscript{\rm 3},
Shenhan Qian\textsuperscript{\rm 4},
Hanling Zhang\textsuperscript{\rm 5}\thanks{Corresponding author.},
Shenghua Gao\textsuperscript{\rm 4,6} \\
\textsuperscript{\rm 1}College of Computer Science and Electronic Engineering, Hunan University \\
\textsuperscript{\rm 2}Manycore Research Institute, Manycore \quad
\textsuperscript{\rm 3}Meituan Group \\
\textsuperscript{\rm 4}ShanghaiTech University \quad
\textsuperscript{\rm 5}School of Design, Hunan University \\
\textsuperscript{\rm 6}Shanghai Engineering Research Center of Intelligent Vision and Imaging \\
{\tt\small \{zyyin, jh\_hlzhang\}@hnu.edu.cn} \quad 
{\tt\small jiajia@qunhemail.com} \quad
{\tt\small luoweixin@meituan.com} \\
{\tt\small \{qianshh, gaoshh\}@shanghaitech.edu.cn}
}

\maketitle

\begin{abstract}
This paper proposes a framework for the interactive video object segmentation (VOS) in the wild where users can choose some frames for annotations iteratively. Then, based on the user annotations, a segmentation algorithm refines the masks. The previous interactive VOS paradigm selects the frame with some worst evaluation metric, and the ground truth is required for calculating the evaluation metric, which is impractical in the testing phase. In contrast, in this paper, we advocate that the frame with the worst evaluation metric may not be exactly the most valuable frame that leads to the most performance improvement across the video. Thus, we formulate the frame selection problem in the interactive VOS as a Markov Decision Process, where an agent is learned to recommend the frame under a deep reinforcement learning framework. The learned agent can automatically determine the most valuable frame, making the interactive setting more practical in the wild. Experimental results on the public datasets show the effectiveness of our learned agent without any changes to the underlying VOS algorithms. Our data, code, and models are available at \url{https://github.com/svip-lab/IVOS-W}.
\end{abstract}

\section{Introduction}

\begin{figure}[t]
    \centering
    \includegraphics[width=\linewidth]{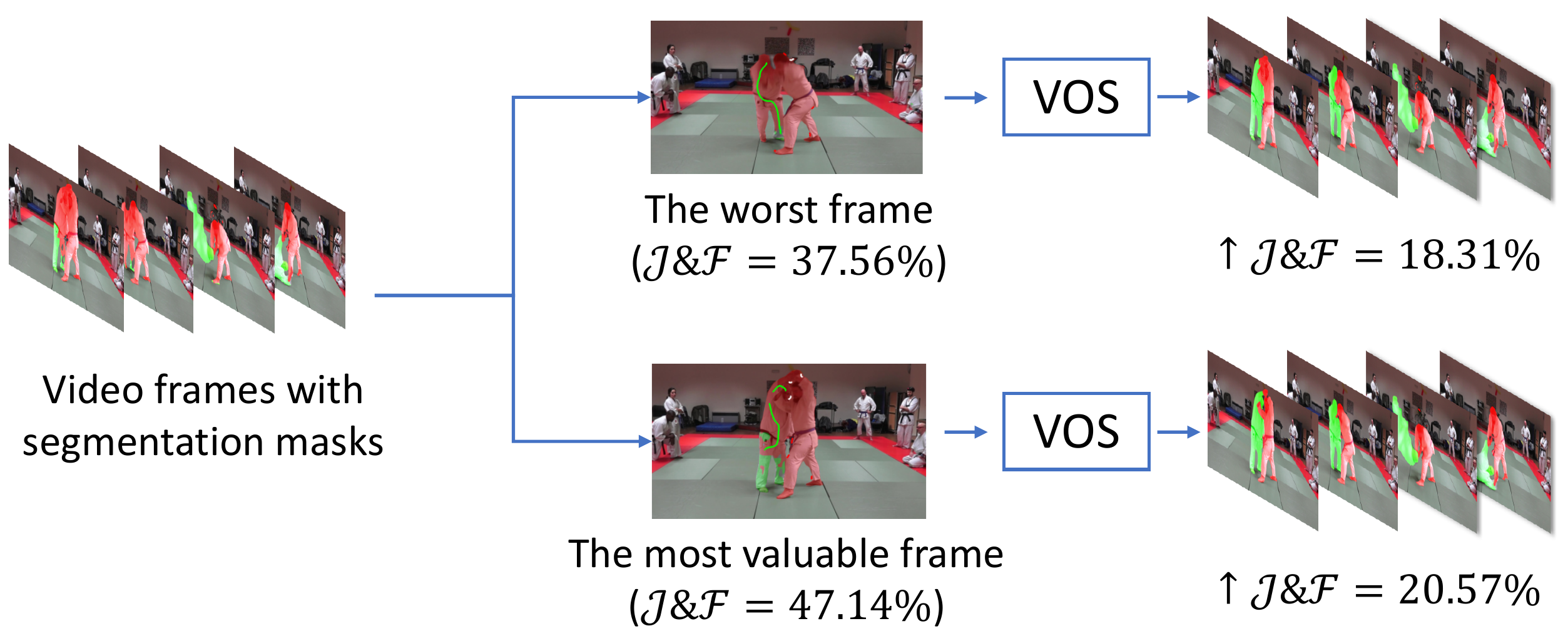}
    \caption{The frame with the worst segmentation quality \vs the most valuable frame in a single round. The frame with the worst segmentation quality only improves the performance of VOS by \SI{18.31}{\percent}, while the most valuable one improves the performance by \SI{20.57}{\percent}.\label{fig:teaser}}
\end{figure}

Video object segmentation aims to segment the objects of interest in a video sequence. It has been widely applied to many downstream applications such as video editing and object tracking. Recently, DAVIS dataset~\cite{PerazziPMVGS16, DAVIS17} and YouTube-VOS dataset~\cite{XuYFYYLPCH18} are introduced and significantly drive forward this task. However, collecting such densely-annotated datasets is expensive and time-consuming. For example, labeling a single object in one frame of DAVIS dataset requires more than \num{100} seconds~\cite{DAVIS18}, finally resulting in either limited sizes~\cite{PerazziPMVGS16, DAVIS17} or coarse annotations~\cite{XuYFYYLPCH18} in the existing VOS datasets.

To minimize the human efforts, \citet{DAVIS18} introduces a human-in-the-loop VOS setting, \ie, the interactive VOS with scribble supervision. Specifically, the interactive VOS algorithm will predict an initial segmentation mask for each frame based on the initial scribbles provided by a user. It will then gradually refine the segmentation masks with additional scribbles of some frames selected by the user, who may evaluate the result by the segmentation quality between the predictions and the ground truths. Whereas, the ground-truth segmentation masks are not available in practice, so the user cannot select a potential frame based on the segmentation quality. Further, the frame with the worst segmentation quality may not be exactly the most valuable one contributing the most to the refinement performance, as shown in \figref{fig:teaser}. In this paper, we claim that \emph{the most valuable frame is not necessarily the one with the worst segmentation quality for the interactive VOS task.}

This paper aims to develop a criterion for determining the \emph{worthiness} of the frame. The worthiness of a frame reflects how much it can improve the segmentation performance across the video sequence if it is selected to provide additional scribbles. However, measuring the worthiness is difficult due to the complexity and variety of videos and uncertainties in the refinement process. To this end, we formulate the frame recommendation problem as a Markov Decision Process (MDP) and train the recommendation agent with Deep Reinforcement Learning (DRL). To narrow the state space, we define the state as the segmentation quality of each frame instead of the image frames and segmentation masks. We also include the recommendation history of each frame in the state. Inspired by Mask Scoring R-CNN~\cite{HuangHGHW19}, we use a segmentation quality assessment module to estimate the segmentation quality. Given the user scribbles on the recommended frame, we leverage the off-the-shelf interactive VOS algorithms~\cite{OhLXK19IPN, MiaoWY20, HeoKK20} to refine the segmentation masks. Without any ground-truth information, the learned agent can recommend the frame. To further evaluate the ability of generalization of our agent, we follow the DAVIS dataset~\cite{DAVIS18} to extend a subset of YouTube-VOS dataset~\cite{XuYFYYLPCH18} with initial scribbles. The experimental results show that our learned agent outperforms all baseline frame selection strategies on DAVIS and YouTube-VOS dataset without any changes to underlying VOS algorithms, whether the ground truth is available or not.

In summary, {\bf our contributions} are as follows: (i) We demonstrate that the frame used in the current interactive VOS paradigm, \ie, the frame with the worst segmentation quality, for user annotation is not the best one. (ii) We propose a novel deep reinforcement recommendation agent for interactive VOS, where the agent recommends the most valuable frame for user annotation. The agent does not require any ground-truth information in the testing phase, therefore it is more practical. (iii) Following the interactive VOS setting~\cite{DAVIS18}, we extend a subset of YouTube-VOS dataset~\cite{XuYFYYLPCH18} with initial scribbles for performance evaluation. (iv) Extensive experiments on the challenging datasets, namely DAVIS dataset and YouTube-VOS dataset, validate the effectiveness of our proposed method.

\section{Related Work}

\subsection{Semi-supervised VOS}

Semi-supervised VOS aims to segment objects based on the object mask given in the first frame. With the advent of deep learning in computer vision, Convolutional Neural Networks (CNNs) have recently been investigated to solve the VOS task. One line of work~\cite{CaellesMPLCV17, ShinRKLSSK17, ChenPTMVG18, HuHS18, ZengLGXFU19, VoigtlaenderCSALC19} detects the objects using the appearance in the first frame. For instance, OSVOS~\cite{CaellesMPLCV17} fine-tunes the network using the first-frame ground truth when testing. FEELVOS~\cite{VoigtlaenderCSALC19} uses pixel-level embedding together with a global and local matching mechanism. Another line of work~\cite{PerazziKBSS17, OhLSJ18, KhorevaBIBS19} learns to propagate the segmentation mask from one frame to the next. DyeNet~\cite{LiC18} takes advantage of both detection and mask propagation approaches. Recently, \citet{GriffinC19} demonstrates that instead of using the first frame as the prior, selecting another frame for annotation will lead to performance improvement. Similar to \cite{GriffinC19}, in this paper, we find this is also applicable to the interactive VOS setting.

\subsection{Interactive VOS}

Interactive VOS relies on the user input, such as scribbles~\cite{DAVIS18, OhLXK19IPN, MiaoWY20, HeoKK20} or points~\cite{ChenPTMVG18}, to segment objects of interest in an interactive manner. \citet{DAVIS18} proposes a CNN-based method built upon OSVOS and fine-tunes the model based on the user annotations in each round. IPN~\cite{OhLXK19IPN} and ATNet~\cite{HeoKK20} use two segmentation networks to handle interaction and propagation, respectively. ATNet~\cite{HeoKK20} further uses a global and local transfer module to transfer segmentation information to other frames. Built upon FEELVOS, MANet~\cite{MiaoWY20} employs a memory aggregation mechanism to record all the previous user annotations. However, all these approaches follow the paradigm of \cite{DAVIS18} and assume the user selects the frame with the worst segmentation quality. In this paper, we argue that \emph{the frame with the worst segmentation quality is not exactly the one with the most potential performance improvement.}

\begin{figure*}[t]
    \centering
    \includegraphics[width=\linewidth]{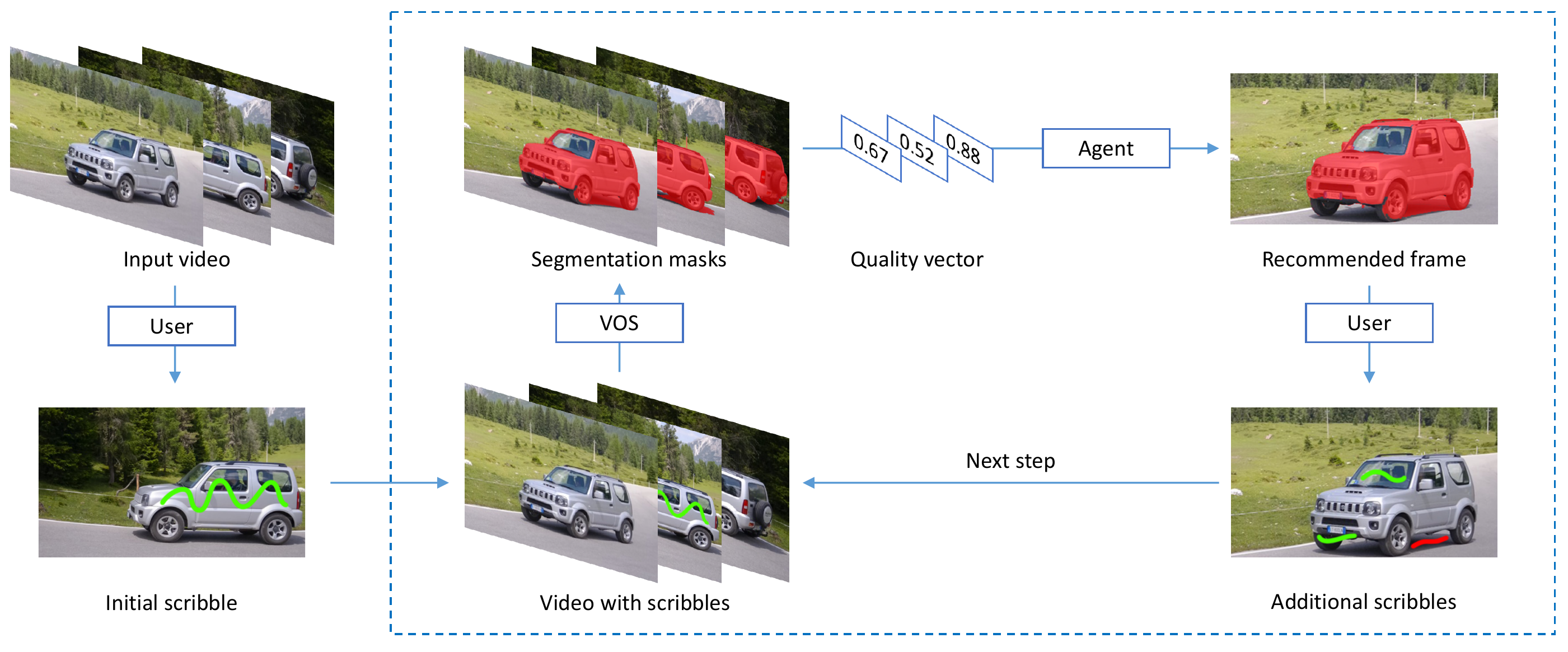}
    \caption{Our proposed interactive VOS framework. In the beginning, the user selects one frame that best represents objects of interest and labels them with initial scribbles. Then, we adopt off-the-shelf VOS algorithm to segment the target object and estimate the segmentation quality for each frame. Afterward, taking the segmentation quality as input, the agent recommends the most valuable frame to the user, who finally draws additional scribbles to refine the masks. Later, the VOS algorithm, the agent, and the user constitute a loop that iteratively refines the predicted masks.\label{fig:pipeline}}
\end{figure*}

\subsection{Reinforcement Learning in Vision and Video}

Reinforcement Learning (RL) is a promising approach to tackle sequential decision-making problems. Many methods try to formulate vision tasks in the spatial and temporal domain as sequential decision-making problems and apply RL to solve them at different levels. \citet{SongMM18} proposes an RL-based method to gradually generate a set of points for the interactive image segmentation. The generated points are used to refine segmentation via an off-the-shelf segmentation algorithm. Some approaches~\cite{RenLWTZ18, RenYLYZ18, ChenWLWL18} introduce RL to tackle the object tracking problem by learning to transfer the bounding box of the target object from the previous frame to an appropriate place in a new frame without scanning all the possible regions. Some other approaches are proposed to locate key-frames in a video sequence for more effective processing at the frame-level. For example, \citet{TangTLLZ18} use RL to find a criterion to select a set of representative frames for action recognition. \citet{WangHW19} learn to locate the activity in a video according to the query language by leveraging an agent to observe selected frames in a video to find the temporal boundaries. \citet{GaoXDSX19} locate the start frame of action in an untrimmed video by conducting a class-agnostic start detector based on observing the action scores for each frame. \citet{HuCHY20} leverage RL to determine a set of most informative frames and group relevant relations inferred from the selected frames.

Recently, RL is also introduced to the VOS. \citet{HanYZCL18} integrates the RL into the VOS task to refine the bounding box when propagating the results from the previous frame to the current frame. \citet{GowdaEHS20} group the object proposals sequentially over both space and time. \citet{Chei19} uses RL to locate a patch of the area as hard attention for VOS to perform the segmentation based on a set of collected memory. \citet{SunXLXF20} propose a method to generate a pixel-level region of interest for more effective online adaptation. Similar to \cite{SunXLXF20}, our proposed method is based on existing VOS methods but focuses on frame-level optimization rather than pixel-level, which is more compatible with the interactive VOS setting.

\section{Methods}

Given $N$ frames $\{ I^1, I^2, \ldots, I^N \}$, the corresponding segmentation masks $\{ M^1, M^2, \ldots, M^N \}$ for the target objects of interest and any previously provided annotations, the agent recommends the most valuable frame for additional user annotation. \figref{fig:pipeline} shows the overall pipeline of our proposed framework.

\subsection{Learning to Recommend}

We formulate the frame recommendation problem in the interactive VOS as an MDP, where the frame selection is only based on the segmentation masks at each step. Specifically, considering the $t$-th iteration, the agent observes the segmentation masks, which are regarded as state $s_t$, and determines the recommendation action $a_t$. Then, the state $s_t$ is transferred to $s_{t+1}$ by the VOS algorithm, and the corresponding reward $r_t$ will be obtained.

\PAR{State.} Intuitively, the state should contain enough information, such as video frames and segmentation masks. However, this leads to higher dimensional state space. Thus, we use the segmentation quality $q_t \in [0, 1]^N$ as a proxy of video frames and masks, where $N$ represents the total number of frames in the video sequence. We further include recommendation history $h_t \in \{0, 1, ..., T\}^N$, where $T$ is the maximum number of interaction, and the $n$-th value in $h_t$ represents the number of times that the $n$-th frame is recommended. Thus, the state $s_t$ is defined as:
\begin{align}
    s_t = \Call{Concat}{q_t, h_t},
\end{align}
where $\Call{Concat}{\cdot}$ denotes the concatenation operator.

\PAR{Action.} The action $a_t \in \{1, \ldots, N\}$ at $t$-th iteration is to determine the next frame for user annotation. We design a Bi-Directional Long Short-Term Memory (LSTM) based network\footnote{Please refer to supplementary material for the detail of network architecture.} to learn the expected recommendation agent. The network takes the state $s_t$ as input, and outputs the action $a_t$. The action of the agent is the recommended frame index.

\PAR{Reward.} The reward reflects the quality of the learned frame recommendation strategy. In the interactive VOS, it is impractical to measure the worthiness of each frame in a single interaction since the contribution of the annotated frame to the final performance cannot be determined without global optimization. Inspired by~\cite{ReinkeUD17}, we design a goal-only reward based on the final performance $P$ achieved by the action sequence until the maximum number of iterations $T$ is reached.

We expect that the learned recommendation policy is at least better than the random selection policy when $t=T$. To get the performance $\hat{P}$ of the random selection policies, we first run experiments \num{30} times with the random selection strategy for each training video sequence. We assume that $\hat{P}$ follows the $t$-distribution, and get the expected mean value $\hat{\mu}$ and variance $\hat{\sigma}$. An intuitive reward function can be designed by comparing the performances between the learned recommendation policy and the random selection policy:
\begin{align}
    r_t^\textrm{goal} (P) = \frac{P - \hat{\mu}}{\hat{\sigma}}.
    \label{eq:reward:naive}
\end{align}

The reward in \eqref{eq:reward:naive} is positive when the performance $P$ is greater than the average performance $\hat{\mu}$ of the random selection policy. Otherwise, the reward will be negative.

In practice, we find that it is not sufficient to make learned agent only better than the average performance of the random selection policy. Thus, we set the reward positive only if $P > \hat{\mu} + \hat{\sigma}$. The final reward is formulated as follows:
\begin{align}
    r_t^\textrm{goal} (P) = \frac{P - (\hat{\mu} + \hat{\sigma})}{\hat{\sigma}}.
    \label{eq:reward:goal}
\end{align}

We set the reward $r_t^\textrm{goal} = 0$ when $t < T$, since the contribution of the intermediate actions to the final performance improvement cannot be measured directly.

Due to the motion and viewpoint, the appearance of the object may change significantly in the video. Intuitively, the recommended frames should cover more distinct frames, which may lead to better performance. To this end, we design an auxiliary reward at each step to encourage more diverse recommendation frames and punish the action that is not the fewest one in the action history:
\begin{equation}
    r_t^\textrm{aux} =
    \begin{cases}
        1,  & a_t = \arg\min h_t, \\
        -1, & \textrm{otherwise.}
    \end{cases}
    \label{eq:reward:aux}
\end{equation}

\PAR{Double Q-learning.} We solve this MDP by the double Q-learning~\cite{HasseltGS16}. Considering both two rewards, the underlying action-value function for the step $t$ is defined as follows:
\begin{align}
    Q^*_{t} =
    \begin{cases}
        \delta \cdot r_t^\textrm{goal}, & t = T, \\
        \delta \cdot r_t^\textrm{aux} + \gamma \cdot Q^T (s_{t+1}, a_{t+1}), & t < T,
    \end{cases}
\end{align}
we set the scaling factor $\delta = 0.1$ and the discount factor $\gamma = 0.95$. We use the policy network $Q^P (\cdot)$ to determine the action by $a_{t+1} = \mathrm{argmax}_a Q^P (s_{t+1}, a)$, and use the target network $Q^T (\cdot)$ to evaluate the value of the action $a_{t+1}$ by $Q^T (s_{t+1}, a_{t+1})$.

We use the mean squared error loss $\Call{MSE}{\cdot}$ to supervise the learning of the agent:
\begin{align}
    \mathcal{L}_\textrm{agent} = \Call{MSE}{Q_t, Q^*_t}.
    \label{eq:loss}
\end{align}

\PAR{Task decomposition.} The standard RL focuses on maximizing the reward received from the whole episode (\eg, actions across the $T$ interactions) and only consider maximizing the final performance. However, the interactive VOS aims to achieve the highest performance with minimal interactions. This motivates us to treat any interaction as an independent procedure and decompose the frame recommendation task with a maximum number of $T$ iterations into $T$ sub-tasks to maximize the performance at each interaction. For each sub-task, the maximum number of iteration $T^\prime = 1, \ldots, T$. Thus, $s_t$ can be intermediate state in the sub-task with $t < T^\prime$. Meanwhile, the $s_t$ is the terminal state for sub-task with $t = T^\prime$.

\subsection{Segmentation Quality Assessment}

To narrow the state space, we use the segmentation quality as a proxy state for our frame recommendation agent. Inspired by Mask Scoring R-CNN~\cite{HuangHGHW19}, we use a quality assessment module to estimate the segmentation quality for each frame.

Suppose that there are $K$ target objects of interest in the video. We first calculate the tight bounding box $B^{n, k}$ containing the foreground mask for each instance $k$ based on the segmentation probability map $\hat{M}^{n, k}$. Then, we enlarge the bounding box $B^{n, k}$ by a factor of \num{1.5}. To ignore the background regions, we crop the RGB image $I^n$ and corresponding probability map $\hat{M}^{n, k}$ based on the enlarged bounding box $B^{n, k}$. Then, we concatenate the cropped RGB image and probability map as the input of the segmentation quality assessment module and obtain the segmentation quality estimation $q^{n, k}$ of each object of interest. We implement this module with a ResNet-50~\cite{HeZRS16} followed by a fully connected layer from \num{2048} to \num{1}.

The segmentation quality $q^n$ of each frame is the average segmentation qualities over all objects of interest within each frame:
\begin{align}
    q^n & = \frac{1}{K} \sum_{k=1}^K q^{n, k}.
\end{align}

We use $\Call{MSE}{\cdot}$ to supervise the learning of the segmentation quality:
\begin{align}
    \mathcal{L}_\textrm{quality} & = \Call{MSE}{q^{n, k}, q^{*, n, k}},
\end{align}
where $q^{*, n, k}$ is the corresponding ground-truth segmentation quality.

\begin{table*}[t]
    \begin{center}
    \sisetup{
        table-number-alignment = center,
        table-auto-round = true,
        table-figures-decimal = 2,
        table-figures-uncertainty = 2,
        detect-weight = true,
        detect-inline-weight = math,
    }
    \begin{tabular}{ll|SSS|SSS}
        \toprule
        \multirow{2}{*}{Setting} & \multirow{2}{*}{Strategy} & \multicolumn{3}{c|}{DAVIS} & \multicolumn{3}{c}{YouTube-VOS} \tabularnewline
        \cmidrule{3-5} \cmidrule{5-8}
        & & {IPN~\cite{OhLXK19IPN}} & {MANet~\cite{MiaoWY20}} & {ATNet~\cite{HeoKK20}} & {IPN~\cite{OhLXK19IPN}} & {MANet~\cite{MiaoWY20}} & {ATNet~\cite{HeoKK20}} \tabularnewline
        \midrule
        \multirow{2}{*}{Oracle} & {Worst} & 48.02 & 70.85 & 73.68 & \bfseries 44.67 & 66.63 & 74.89 \tabularnewline
        & {Ours} & \bfseries 48.25 & \bfseries 71.11 & \bfseries 74.0144 & 43.86 & \bfseries 66.90 & \bfseries 75.37 \tabularnewline
        \midrule
        \multirow{4}{*}{Wild} & {Random} & 47.52 \pm 0.04 & 69.81 \pm 0.01 & 72.99 \pm 0.03 & 43.22 \pm 0.05 & 65.52 \pm 0.08 & 74.11 \pm 0.08 \tabularnewline
        & {Linspace}  & 46.97 & 70.10 & 72.93 & 42.75 & 65.35 & 73.47 \tabularnewline
        & {Worst} & 47.2568 & 69.3218 & 73.3266 & 43.2898 & 65.9779 & 74.6877 \tabularnewline
        & {Ours} & \bfseries 47.99 & \bfseries 70.82 & \bfseries 74.1047 & \bfseries 43.69 & \bfseries 66.85 & \bfseries 75.33 \tabularnewline
        \bottomrule
    \end{tabular}
    \caption{Quantitative results (AUC) of the interactive VOS on DAVIS and YouTube-VOS dataset.\label{tab:results}}
    \end{center}
\end{table*}

\subsection{Training and Inference}

\PARbegin{Training.} We train the frame recommendation agent on DAVIS dataset. We adopt the VOS algorithm, \ie, ATNet~\cite{HeoKK20}, as the state transition function. It is impractical to train the agent using the whole video sequence due to the various sequence lengths. For each original training sequence, we sample $N^\prime$ consecutive frames to form the sub-sequence and use the corresponding ground-truth segmentation quality to form the state. We use the experience replay mechanism~\cite{SchaulQAS16} to make the training process more stable. At the beginning of the agent training, we fill the experience buffer by randomly selecting the frames and then train the agent with $\epsilon$-greedy policy. To train the segmentation quality assessment module, we use the segmentation masks generated by ATNet.

\PAR{Inference.} Given a test video sequence and initial segmentation masks, the segmentation quality assessment module first estimates the segmentation quality for each frame. Then, the agent takes the segmentation quality and recommendation history as input and outputs $Q$ value for each frame. Finally, we recommend the frame with the highest $Q$ value for user annotation. During testing, we use the whole video sequence.

\section{Experiments}

In this section, we conduct experiments to evaluate the performance of the proposed method on DAVIS dataset~\cite{DAVIS17} and YouTube-VOS dataset~\cite{XuYFYYLPCH18}.

\subsection{Dataset and Evaluation Metrics}

\PARbegin{Datasets.} DAVIS dataset~\cite{DAVIS17} contains \num{60} training sequences and \num{30} validate sequences. DAVIS dataset provides high-quality densely-annotated segmentation mask annotation for each frame. To test the generalization of the proposed method, we further sample \num{50} sequences from YouTube-VOS dataset~\cite{XuYFYYLPCH18}. Since YouTube-VOS dataset does not contain the applicable annotations for the interactive VOS task, we extend such initial scribbles by following~\cite{DAVIS18}.

\PAR{Evaluation metrics.} To validate the performance of segmentation, we use the region similarity in terms of intersection over union $\mathcal{J}$ and the boundary accuracy $\mathcal{F}$ as used in~\cite{PerazziPMVGS16}. \citet{DAVIS18} propose to use the curve of $\mathcal{J} \& \mathcal{F}$ versus the time. Since we focus on evaluating the frame selection strategy, we do not take the time into account. Instead, we use the curve of $\mathcal{J} \& \mathcal{F}$ versus the number of interactions and measure its Area Under Curve (AUC) to validate the interactive setting. The segmentation quality is also measured by $\mathcal{J} \& \mathcal{F}$.

\subsection{Implementation Details}

We implement our model with PyTorch~\cite{PyTorch} and train all models on a single NVIDIA Tesla V100 GPU device. We use Adam~\cite{KingmaB15} optimizer with learning rate \num{5e-6} and batch size \num{32}. The experience buffer is set to \num{5760}. $\epsilon$ decreases from \numrange{0.7}{0.25} over \num{5000} steps exponentially. To accelerate the training process, we set the maximum of interactions $T = 5$ during training, and $T = 8$ during testing following~\cite{DAVIS18}. It is impractical to generate the scribble annotations by human annotators during training, so we use the human-simulated scribbles\footnote{\url{https://github.com/albertomontesg/davis-interactive}} by comparing the segmentation predictions and corresponding ground truths. We set $N^\prime = 25$ to the length of the shortest sequence in the training set. It takes approximately \num{10} hours for the agent to converge.

\subsection{Main Results}

\begin{figure*}[t]
    \centering
    \setlength{\tabcolsep}{1pt}
    \subfloat[IPN~\cite{OhLXK19IPN}]{
        \includegraphics[height=1.4in]{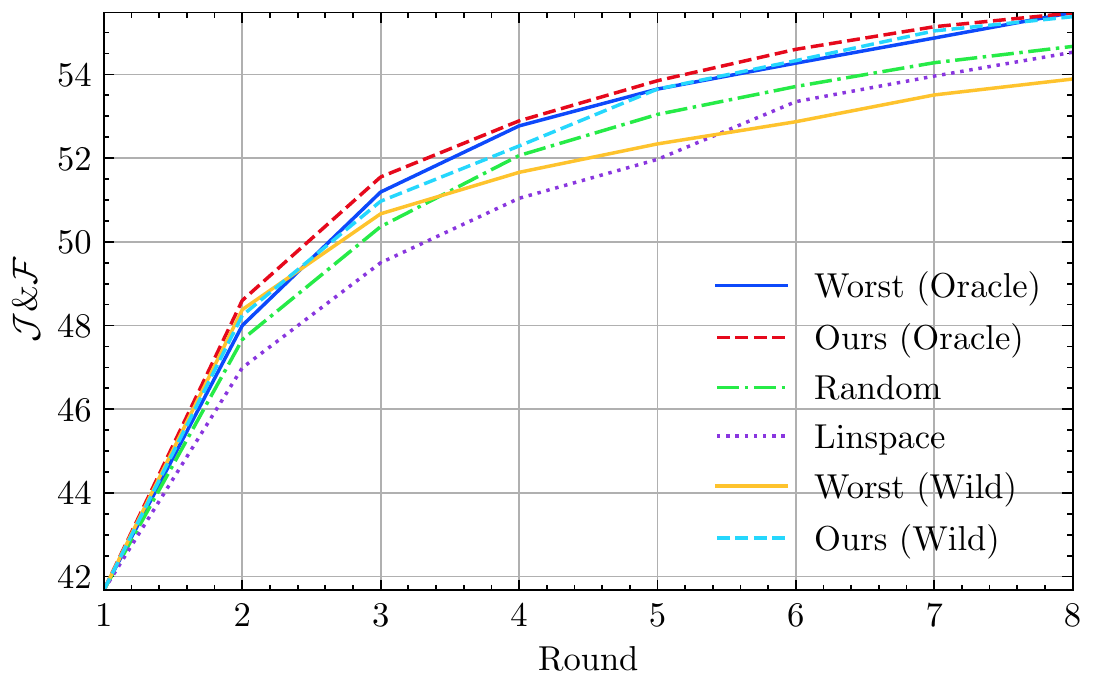}
    }
    \subfloat[MANet~\cite{MiaoWY20}]{
        \includegraphics[height=1.4in]{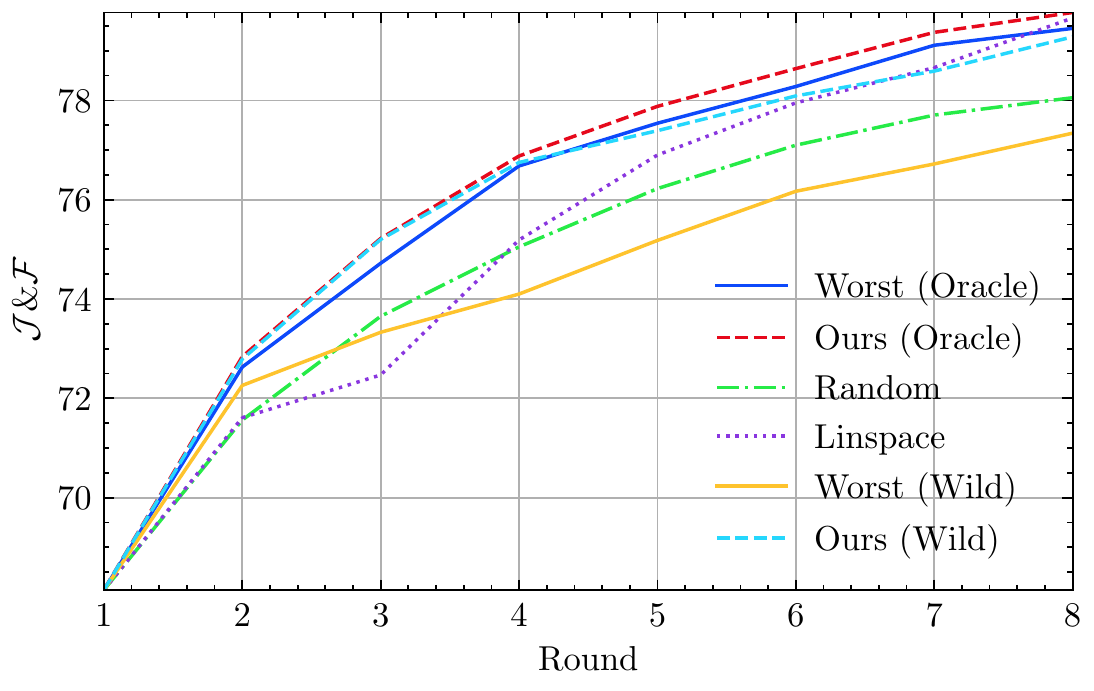}
    }
    \subfloat[ATNet~\cite{HeoKK20}]{
        \includegraphics[height=1.4in]{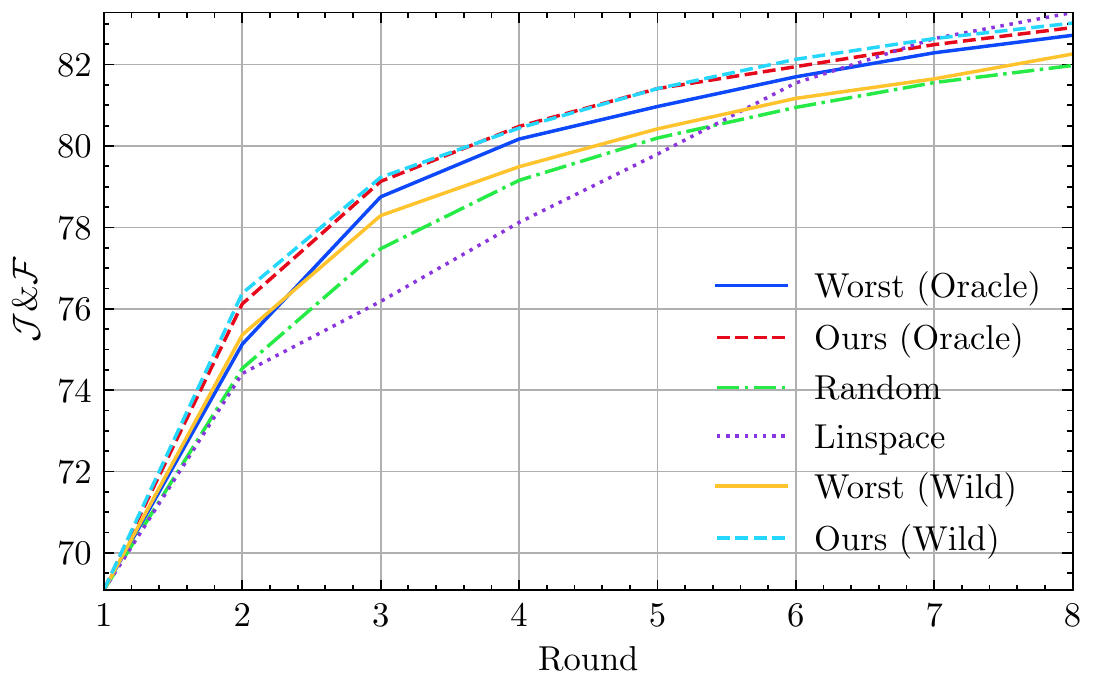}
    }
    \caption{The curve of $\mathcal{J} \& \mathcal{F}$ versus the number of rounds on DAVIS dataset.\label{fig:curves}}
\end{figure*}

\begin{figure*}[t]
    \small
    \centering
    \setlength{\tabcolsep}{1pt}
    \begin{tabular}{ccccccc}
        \includegraphics[width=0.135\linewidth]{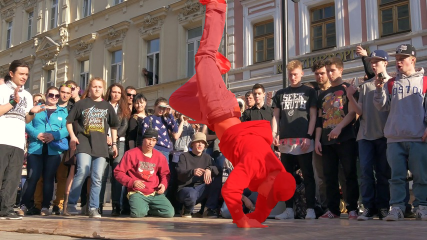} &
        \includegraphics[width=0.135\linewidth]{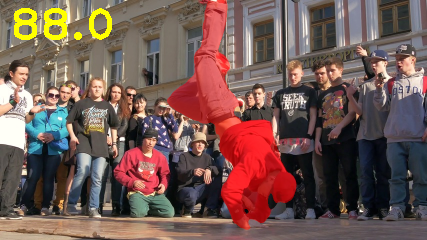} & 
        \includegraphics[width=0.135\linewidth]{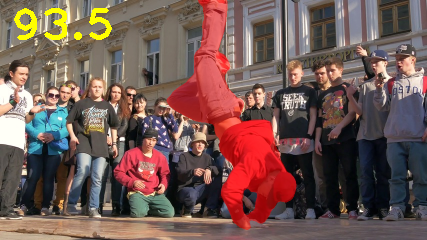} & 
        \includegraphics[width=0.135\linewidth]{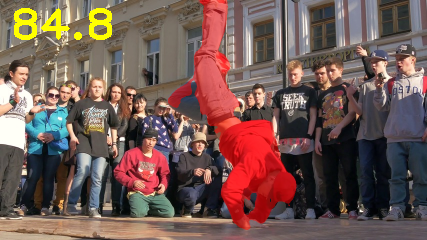} &
        \includegraphics[width=0.135\linewidth]{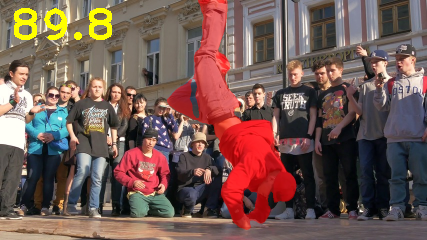} &
        \includegraphics[width=0.135\linewidth]{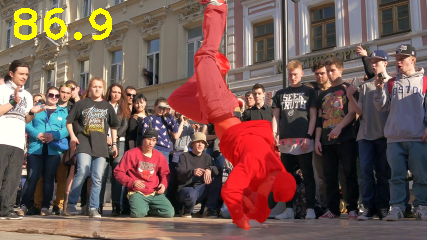} &
        \includegraphics[width=0.135\linewidth]{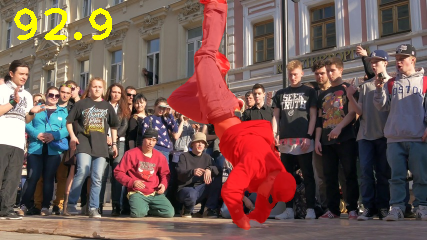}
        \tabularnewline
        \includegraphics[width=0.135\linewidth]{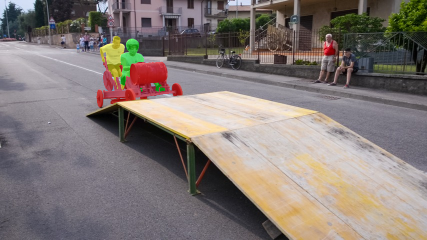} &
        \includegraphics[width=0.135\linewidth]{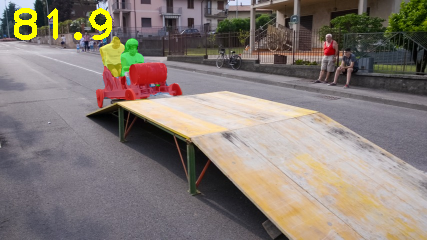} &
        \includegraphics[width=0.135\linewidth]{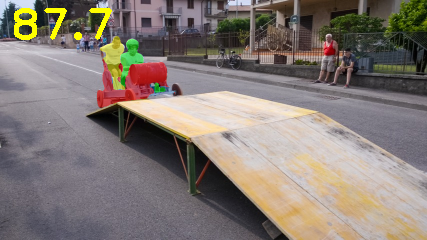} &
        \includegraphics[width=0.135\linewidth]{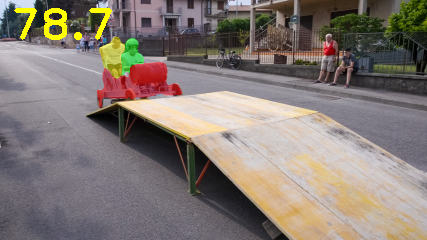} &
        \includegraphics[width=0.135\linewidth]{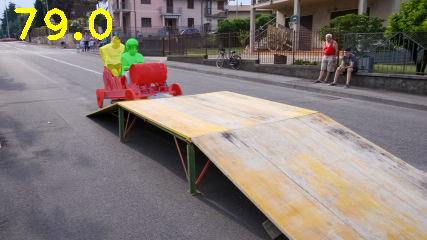} &
        \includegraphics[width=0.135\linewidth]{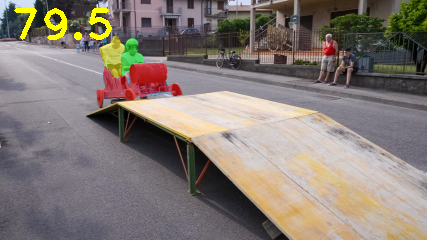} &
        \includegraphics[width=0.135\linewidth]{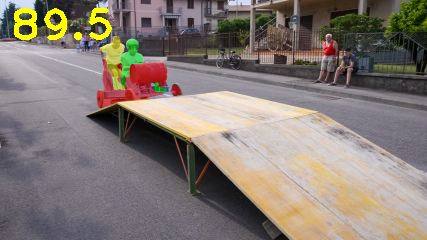}
        \tabularnewline
        \includegraphics[width=0.135\linewidth]{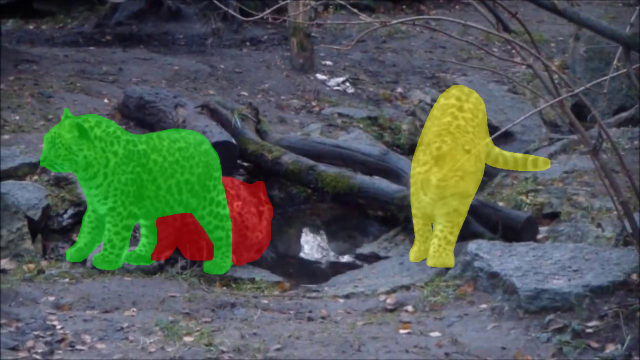} &
        \includegraphics[width=0.135\linewidth]{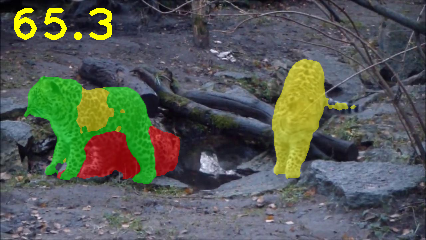} & 
        \includegraphics[width=0.135\linewidth]{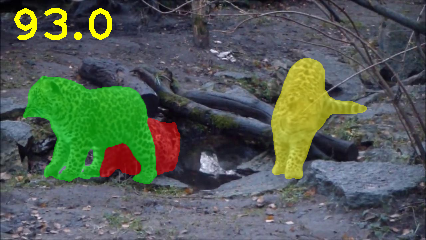} & 
        \includegraphics[width=0.135\linewidth]{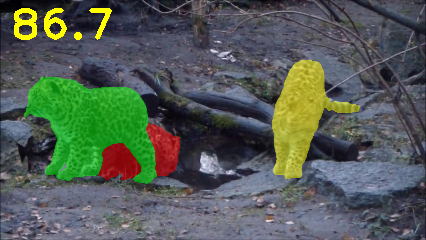} &
        \includegraphics[width=0.135\linewidth]{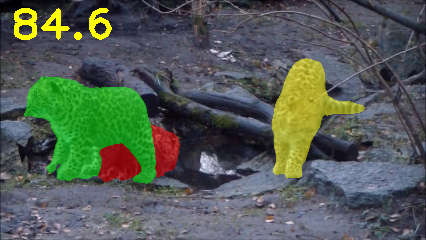} &
        \includegraphics[width=0.135\linewidth]{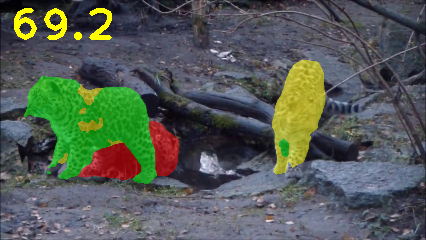} &
        \includegraphics[width=0.135\linewidth]{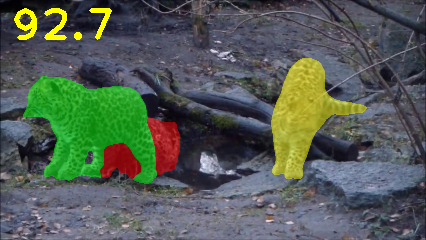}
        \tabularnewline
        \includegraphics[width=0.135\linewidth]{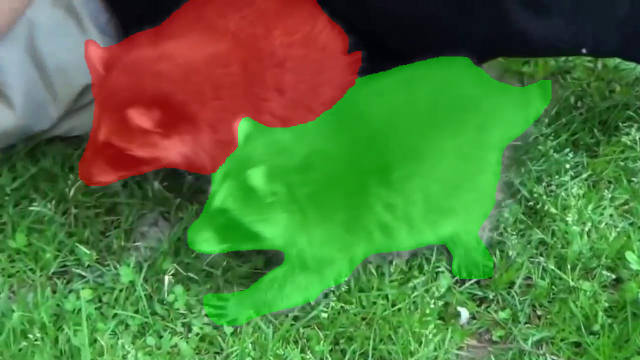} &
        \includegraphics[width=0.135\linewidth]{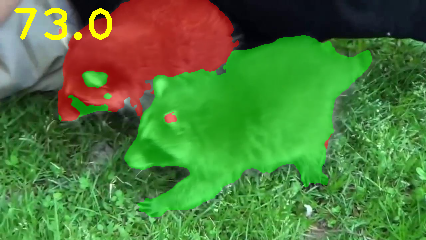} & 
        \includegraphics[width=0.135\linewidth]{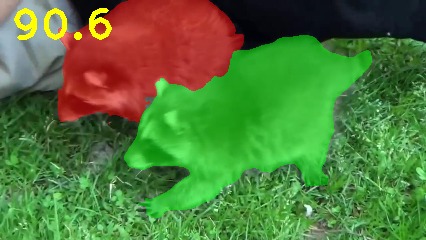} & 
        \includegraphics[width=0.135\linewidth]{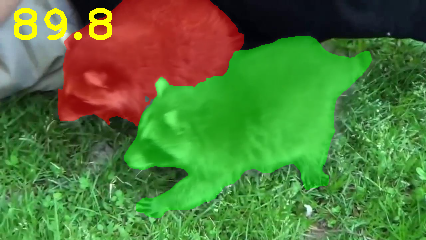} &
        \includegraphics[width=0.135\linewidth]{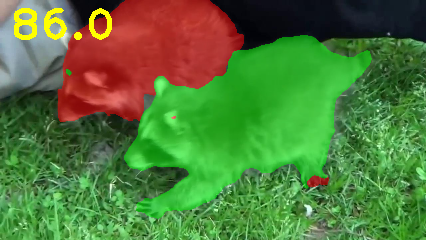} &
        \includegraphics[width=0.135\linewidth]{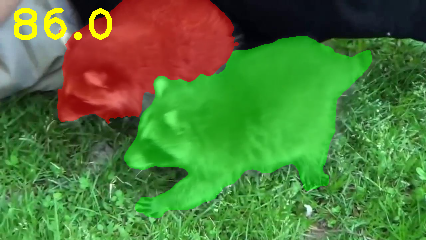} &
        \includegraphics[width=0.135\linewidth]{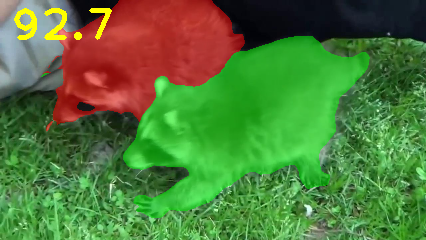}
        \tabularnewline
        Ground-truth & Worst (Oracle) & Ours (Oracle) & Random & Linspace & Worst (Wild) & Ours (Wild) 
    \end{tabular}
    \caption{Qualitative comparison on DAVIS (the first two rows) and YouTube-VOS dataset (the last two rows). All result masks are sampled after 8 rounds. The ground truth is available (``Oracle'') in the second and third columns, while the ground truth is unknown (``Wild'') in the last four columns. We show the segmentation quality $\mathcal{J} \& \mathcal{F}$ on each frame.\label{fig:results}}
\end{figure*}

\PARbegin{Strategies for comparison.} We compare our learned agent under two settings:
\begin{itemize}
    \item {\bf ``Oracle''}: When the ground-truth segmentation mask is available, we compare our method with \cite{DAVIS18}, \ie, select the frame with the worst segmentation quality (``Worst''). In this setting, we feed the ground-truth segmentation quality to our agent.

    \item {\bf ``Wild''}: When the ground-truth segmentation mask is unavailable, we compare our agent with the following frame selection strategies: (i) select uniformly from all frames (``Random''), (ii) select frames with a fixed-length step (``Linspace''), (iii) ``Worst''. In this setting, we use the predicted segmentation quality for ``Worst''. We run ``Random'' selection strategy \num{5} times and report the mean and variance.
\end{itemize}

\begin{figure*}[t]
    \small
    \centering
    \setlength{\tabcolsep}{1pt}
    \begin{tabular}{ccccccccc}
        & Round 1 & Round 2 & Round 3 & Round 4 & Round 5 & Round 6 & Round 7 & Round 8
		\tabularnewline
        \midrule
        \begin{turn}{90}{\hspace{0.5em}{Worst}}\end{turn} &
        \includegraphics[width=0.115\linewidth]{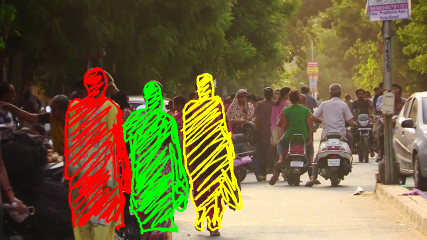} &
        \includegraphics[width=0.115\linewidth]{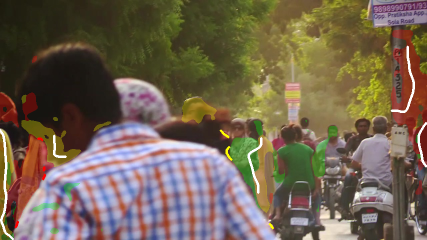} &
        \includegraphics[width=0.115\linewidth]{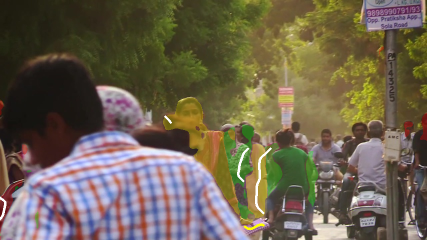} &
        \includegraphics[width=0.115\linewidth]{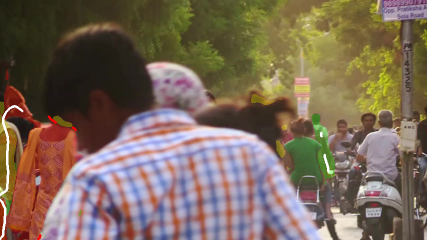} &
        \includegraphics[width=0.115\linewidth]{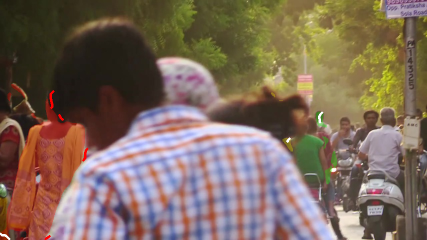} &
        \includegraphics[width=0.115\linewidth]{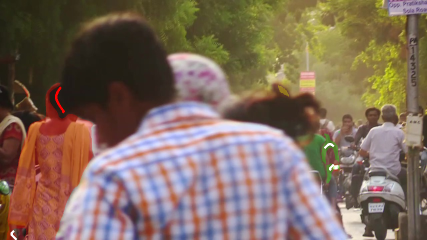} &
        \includegraphics[width=0.115\linewidth]{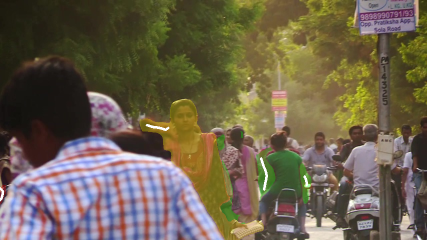} &
        \includegraphics[width=0.115\linewidth]{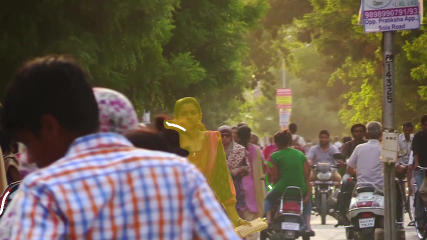}
		\tabularnewline
        & Frame 1 & Frame 59 & Frame 62 & Frame 55 & Frame 54 & Frame 53 & Frame 65 & Frame 63
		\tabularnewline
        \midrule
		\begin{turn}{90}{\hspace{0.75em}{Ours}}\end{turn} & 
        \includegraphics[width=0.115\linewidth]{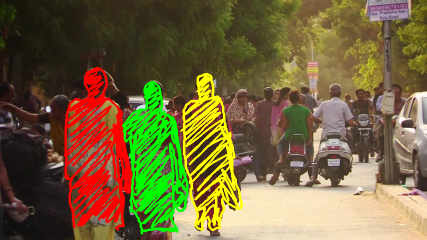} &
        \includegraphics[width=0.115\linewidth]{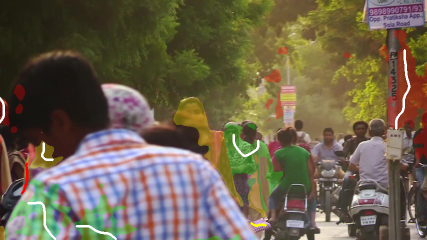} &
        \includegraphics[width=0.115\linewidth]{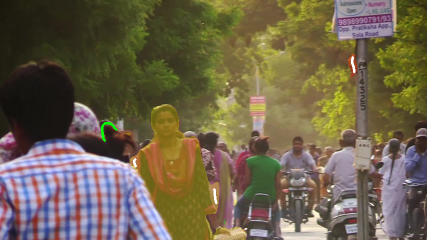} &
        \includegraphics[width=0.115\linewidth]{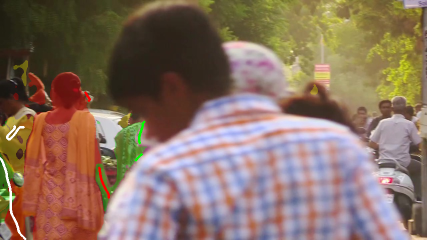} &
        \includegraphics[width=0.115\linewidth]{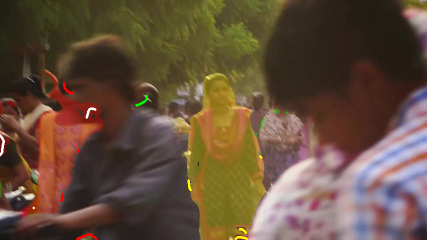} &
        \includegraphics[width=0.115\linewidth]{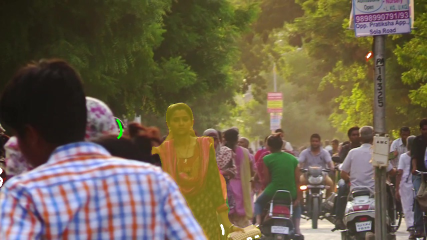} &
        \includegraphics[width=0.115\linewidth]{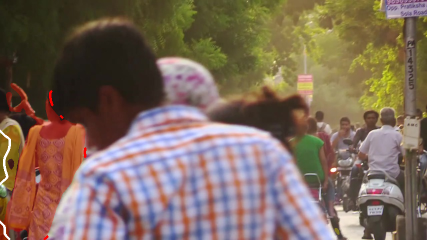} &
        \includegraphics[width=0.115\linewidth]{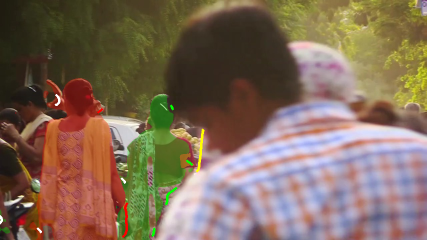} 
		\tabularnewline
		& Frame 1 & Frame 61 & Frame 74 & Frame 49 & Frame 38 & Frame 67 & Frame 54 & Frame 45
        \tabularnewline
    \end{tabular}
    \caption{Recommended frames for the ``india'' sequence in DAVIS dataset. The worst frame selection strategy in the top row achieves \SI{66.92}{\percent} in terms of $\mathcal{J} \& \mathcal{F}$, while ours in the bottom row achieves \SI{72.25}{\percent}.\label{fig:frames}}
\end{figure*}

\PAR{Segmentation algorithm.} We choose three off-the-shelf interactive VOS approaches, IPN~\cite{OhLXK19IPN}\footnote{\url{https://github.com/seoungwugoh/ivs-demo}}, MANet~\cite{MiaoWY20}\footnote{\url{https://github.com/lightas/CVPR2020_MANet}} and ATNet~\cite{HeoKK20}\footnote{\url{https://github.com/yuk6heo/IVOS-ATNet}}, based on their performance and source code availability. All the segmentation algorithms are only trained on DAVIS dataset.

\PAR{Quantitative evaluation.} \tabref{tab:results} shows the quantitative results on DAVIS dataset and YouTube-VOS dataset. We make the following observations: (i) Our learned agent achieves state-of-the-art performance on DAVIS dataset and generalizes well to YouTube-VOS dataset without any changes to the underlying VOS algorithms, no matter if the ground truth is available or not. (ii) Our agent outperforms the worst frame selection strategy (with the exception of IPN) when ground truth is available (``Oracle''), demonstrating that the frame with the worst evaluation result is not exactly the best one for user annotation. (iii) When ground truth is not available (``Wild''), our method also outperforms all baseline strategies. Due to the space limitation, we refer readers to the supplementary material for the curves of all results on YouTube-VOS dataset.

\figref{fig:curves} shows the curves of $\mathcal{J} \& \mathcal{F}$ versus the number of rounds on DAVIS dataset. As one can see, our agent outperforms other frame selection strategies in all rounds when ground truth is available (``Oracle''). When ground truth is not available (``Wild''), our agent can outperform all baselines, \ie, Random, Worst, and Linspace.

\PAR{Qualitative evaluation.} \figref{fig:results} shows the qualitative results of ATNet on DAVIS validation set. We sample results generated by the different frame selection policies after \num{8} rounds. As one can see, our approach produces accurate segmentation masks. We also show the frames recommended by our agent and the worst frames at each round in \figref{fig:frames}. The worst frame selection strategy tends to select a small range of frames. However, the user could not provide additional information for the objects on these frames. We refer readers to supplementary materials for more qualitative results.

\begin{table}[t]
    \centering
    \sisetup{
        table-number-alignment = center,
        table-auto-round = true,
        table-figures-decimal = 2,
        detect-weight = true,
        detect-inline-weight = math
    }
    \begin{tabular}{lSS}
        \toprule
        {Annotator} & {AUC} & {Time (\si{\second})} \tabularnewline
        \midrule
        {Human} & 73.09 & 14.01 \tabularnewline
        {Ours} & \bfseries 74.1047 & \bfseries 0.70 \tabularnewline
        \bottomrule
    \end{tabular}
    \caption{Comparison with humans on DAVIS dataset.\label{tab:human}}
\end{table}

\PAR{Comparison with humans.} We further compare our proposed frame recommendation agent with the human on DAVIS validation set. In this experiment, we adopt ATNet as the VOS algorithm. We overlay the segmentation mask on the RGB image and show the overlaid frame to the human. We only ask the human to select the valuable frame for annotation, and then the chosen frame is annotated by the human-simulated scribbles~\cite{DAVIS18}. As shown in \tabref{tab:human}, our learned agent outperforms the human in both accuracy and efficiency.

\begin{table}[t]
    \centering
    \sisetup{
        table-number-alignment = center,
        table-auto-round = true,
        table-figures-decimal = 2
    }
    \begin{tabular}{lSS}
        \toprule
        {VOS} & {PCC} & {MSE} \tabularnewline
        \midrule
        {IPN~\cite{OhLXK19IPN}} & 0.47 & 0.0470 \tabularnewline
        {MANet~\cite{MiaoWY20}} & 0.42 & 0.0132 \tabularnewline
        {ATNet~\cite{HeoKK20}} & 0.51 & 0.0096 \tabularnewline
        \bottomrule
    \end{tabular}
    \caption{Quantitative results of the segmentation quality assessment module on DAVIS dataset.\label{tab:quality}}
\end{table}

\PAR{Evaluation of the segmentation quality assessment.} To validate the accuracy of the segmentation quality assessment module, we adopt the Pearson correlation coefficient (PCC) and MSE between predictions and their ground truths. As shown in \tabref{tab:quality}, the regression model trained only on the data generated by ATNet can also generalize well to other VOS algorithms.

\begin{figure*}[t]
    \centering
    \setlength{\tabcolsep}{1pt}
    \subfloat[{The reward curve using \eqref{eq:reward:naive}.\label{fig:reward:naive}}]{
        \includegraphics[height=1.5in]{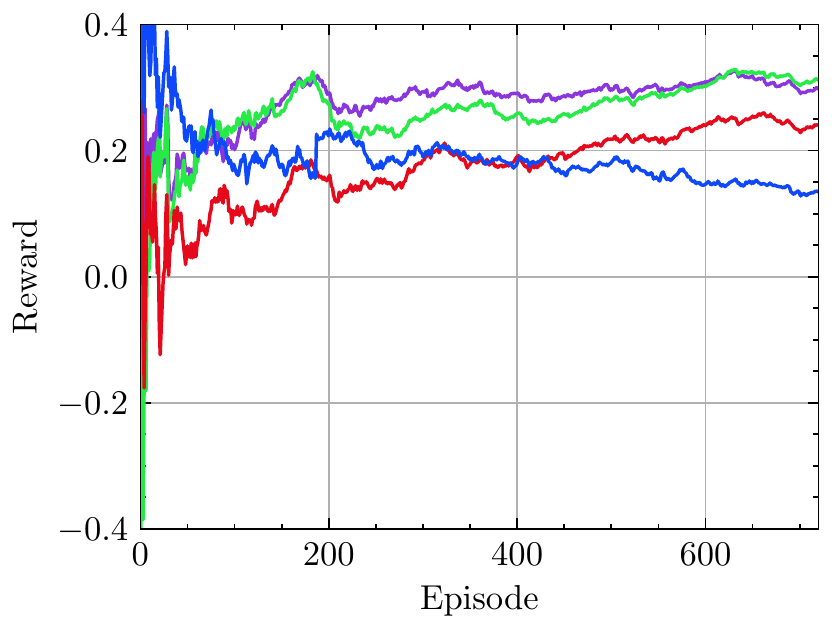}
    }
    \subfloat[{The reward curve using \eqref{eq:reward:goal}.\label{fig:reward:final}}]{
        \includegraphics[height=1.5in]{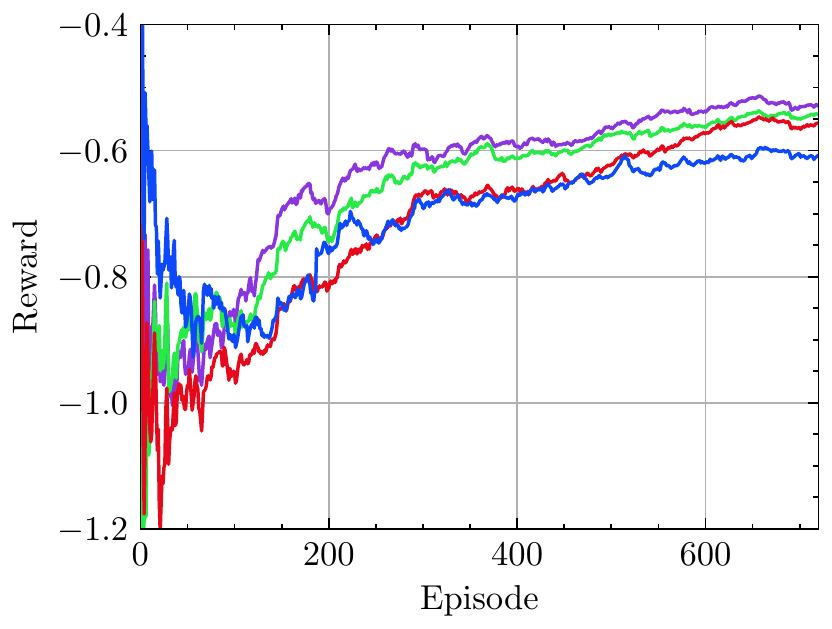}
    }
    \subfloat[{The performance curve.\label{fig:reward:all}}]{
        \includegraphics[height=1.5in]{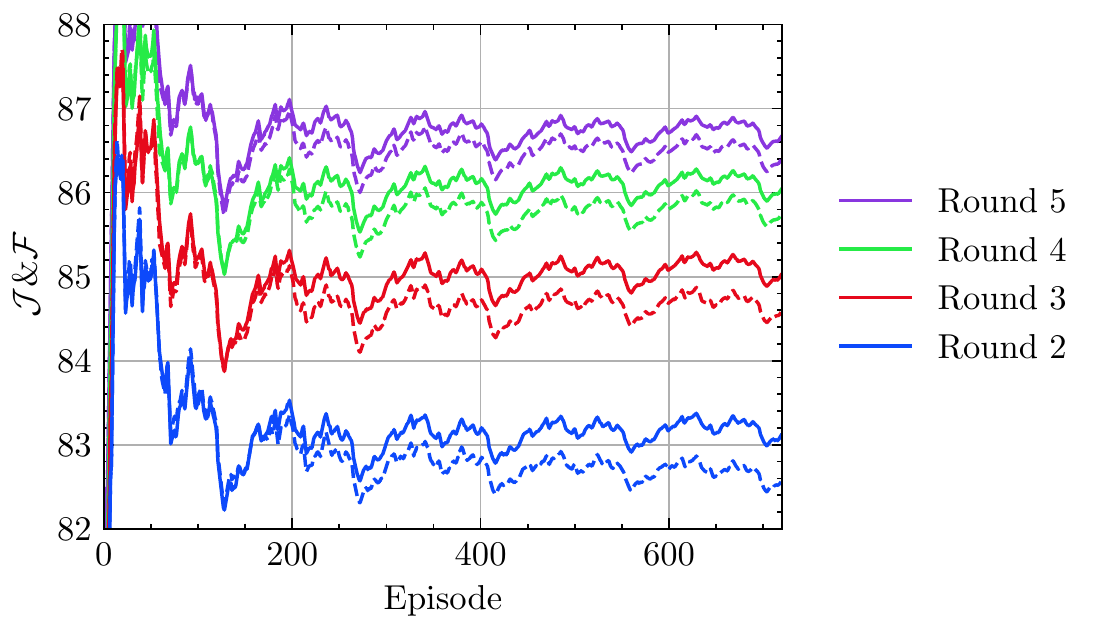}
    }
    \caption{Training curves on DAVIS dataset. (a) and (b) show the reward obtained in each round with the reward function in \eqref{eq:reward:naive} and \eqref{eq:reward:goal}, respectively. (c) shows the segmentation quality in each round. The dashed and solid lines in (c) represent the performance based on \eqref{eq:reward:naive} and \eqref{eq:reward:goal}, respectively. All curves are smoothed using a weighed moving average algorithm.\label{fig:reward}}
\end{figure*}

\begin{figure*}
    \small
    \centering
    \setlength{\tabcolsep}{1pt}
    \begin{tabular}{ccccccccc}
        & Round 1 & Round 2 & Round 3 & Round 4 & Round 5 & Round 6 & Round 7 & Round 8
        \tabularnewline
        \midrule
        \begin{turn}{90}{\hspace{0.5em}{Worst}}\end{turn} &
        \includegraphics[width=0.115\linewidth]{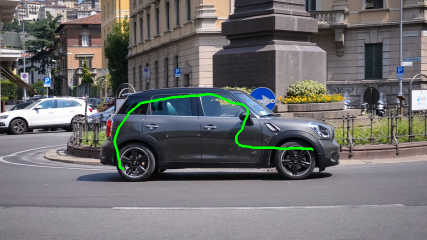} &
        \includegraphics[width=0.115\linewidth]{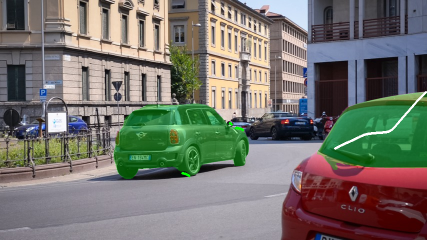} &
        \includegraphics[width=0.115\linewidth]{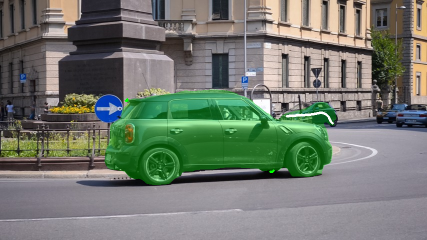} &
        \includegraphics[width=0.115\linewidth]{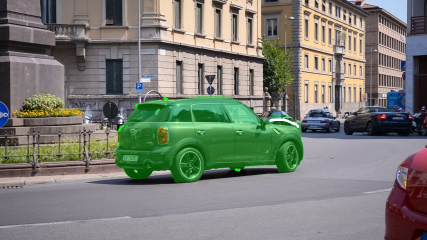} &
        \includegraphics[width=0.115\linewidth]{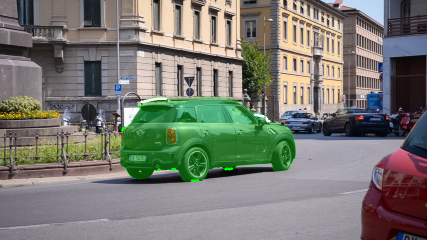} &
        \includegraphics[width=0.115\linewidth]{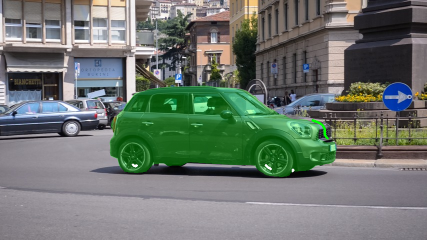} &
        \includegraphics[width=0.115\linewidth]{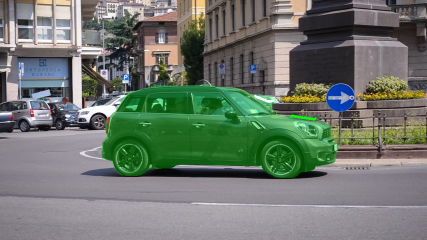} &
        \includegraphics[width=0.115\linewidth]{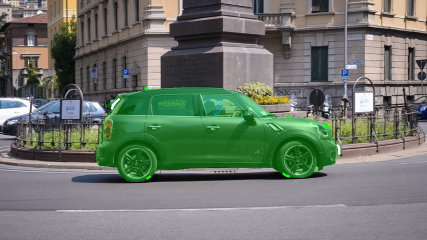}
        \tabularnewline
        & Frame 22 & Frame 74 & Frame 41 & Frame 57 & Frame 61 & Frame 5 & Frame 12 & Frame 28
        \tabularnewline
        \midrule
        \begin{turn}{90}{\hspace{0.75em}{Ours}}\end{turn} & 
        \includegraphics[width=0.115\linewidth]{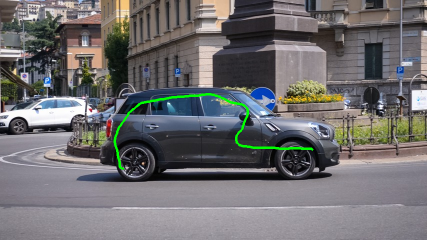} &
        \includegraphics[width=0.115\linewidth]{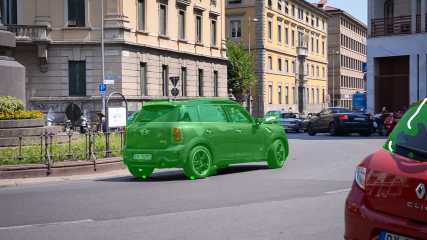} &
        \includegraphics[width=0.115\linewidth]{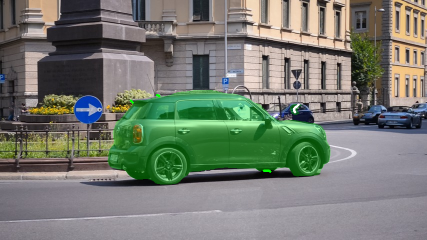} &
        \includegraphics[width=0.115\linewidth]{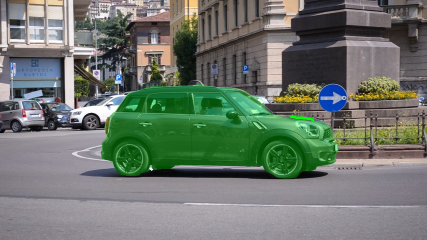} &
        \includegraphics[width=0.115\linewidth]{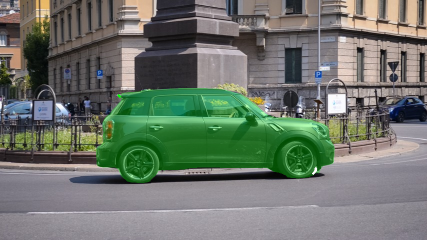} &
        \includegraphics[width=0.115\linewidth]{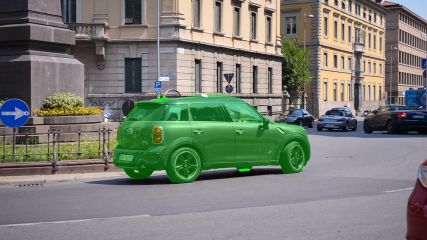} &
        \includegraphics[width=0.115\linewidth]{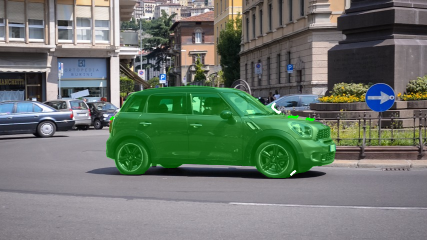} &
        \includegraphics[width=0.115\linewidth]{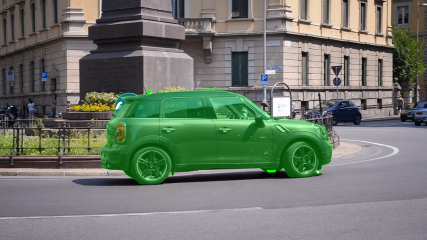} 
        \tabularnewline
        & Frame 22 & Frame 64 & Frame 44 & Frame 13 & Frame 31 & Frame 54 & Frame 07 & Frame 38
        \tabularnewline
    \end{tabular}
    \caption{Failure case. We show the recommended frames for the ``car-roundabout'' sequence in DAVIS dataset. The worst frame selection strategy in the top row achieves \SI{96.75}{\percent} in terms of $\mathcal{J} \& \mathcal{F}$, while ours in the bottom row achieves \SI{94.43}{\percent}.\label{fig:failure}}
\end{figure*}

\subsection{Ablation Studies}

We run several ablation studies to analyze the frame recommendation agent. In all ablation studies, we adopt the ATNet as the VOS algorithm and report the AUC on DAVIS validation set.

\begin{table}[t]
    \centering
    \sisetup{
        table-number-alignment = center,
        table-auto-round = true,
        table-figures-decimal = 2,
        detect-weight = true,
        detect-inline-weight = math
    }
    \begin{tabular}{lSS}
        \toprule
        {Variants} & {Oracle} & {Wild} \tabularnewline
        \midrule
        {\eqref{eq:reward:naive}} & 71.8211 & 71.9709 \tabularnewline
        {\eqref{eq:reward:goal}} & \bfseries 74.01 & \bfseries 74.1047 \tabularnewline
        \bottomrule
    \end{tabular}
    \caption{Reward function.\label{tab:ab:function}}
\end{table}

\PAR{Reward function.} We first verify the effectiveness of the proposed reward function. We compare the two reward functions, \ie, \eqref{eq:reward:naive} and \eqref{eq:reward:goal}. The reward in \eqref{eq:reward:naive} is positive if $P > \hat{\mu}$, while the reward in \eqref{eq:reward:goal} is positive if $P > \hat{\mu} + \hat{\sigma}$. The results are shown in \tabref{tab:ab:function}. As expected, the proposed reward function has better results.

We further show the change in reward according to the training episode from \num{2}nd round to \num{5}th round in \figref{fig:reward}. As shown in \figref{fig:reward:naive}, the reward in \eqref{eq:reward:naive} is mostly positive, and the performance is hard to improve after a certain training episode. In contrast, the reward in \eqref{eq:reward:goal} shown in \figref{fig:reward:final} can continuously improve. \figref{fig:reward:all} shows the change in $\mathcal{J} \& \mathcal{F}$ of the training process. As expected, the reward in \eqref{eq:reward:goal} has better results than that in \eqref{eq:reward:naive}.

\begin{table}[t]
    \centering
    \sisetup{
        table-number-alignment = center,
        table-auto-round = true,
        table-figures-decimal = 2,
        detect-weight = true,
        detect-inline-weight = math
    }
    \begin{tabular}{lSS}
        \toprule
        {Variants} & {Oracle} & {Wild} \tabularnewline
        \midrule
        {$r^\textrm{goal}$} & 73.7465 & 73.7577 \tabularnewline
        {$r^\textrm{aux}$} & 72.0605 & 72.0009 \tabularnewline
        {$r^\textrm{goal} + r^\textrm{aux}$} & \bfseries 74.01 & \bfseries 74.1047 \tabularnewline
        \bottomrule
    \end{tabular}
    \caption{Reward.\label{tab:ab:reward}}
\end{table}

\PAR{Reward.} To evaluate the effectiveness of goal-only reward $r^\textrm{goal}$ in \eqref{eq:reward:goal} and auxiliary reward $r^\textrm{aux}$ in \eqref{eq:reward:aux}, we remove either one of them. The results are shown in \tabref{tab:ab:reward}. As one can see, the agent trained with both rewards achieves the best performance. This demonstrates that both segmentation quality and frame selection diversity are helpful to the frame recommendation.

\begin{table}[t]
    \centering
    \sisetup{
        table-number-alignment = center,
        table-auto-round = true,
        table-figures-decimal = 2,
        detect-weight = true,
        detect-inline-weight = math
    }
    \begin{tabular}{lSS}
        \toprule
        {Variants} & {Oracle} & {Wild} \tabularnewline
        \midrule
        {$q_t$} & 73.7088 & 73.5501 \tabularnewline
        {$h_t$} & 73.9151 & 73.9151 \tabularnewline
        {$q_t+h_t$} & \bfseries 74.01 & \bfseries 74.1047  \tabularnewline
        \bottomrule
    \end{tabular}
    \caption{State.\label{tab:ab:state}}
\end{table}

\PAR{State.} To evaluate the effectiveness of the state, we remove either the segmentation quality $q_t$ or the recommendation history $h_t$. As shown in \tabref{tab:ab:state}, both states play an important role in representing the agent.

\begin{table}[t]
    \centering
    \sisetup{
        table-number-alignment = center,
        table-auto-round = true,
        table-figures-decimal = 2,
        detect-weight = true,
        detect-inline-weight = math
    }
    \begin{tabular}{lSS}
        \toprule
        {Variants} & {Oracle} & {Wild} \tabularnewline
        \midrule
        {\xmark} & 72.5831 & 72.5535 \tabularnewline
        {\cmark} & \bfseries 74.01 & \bfseries 74.1047 \tabularnewline
        \bottomrule
    \end{tabular}
    \caption{Task decomposition.\label{tab:ab:task}}
\end{table}

\PAR{Task decomposition.} Finally, we investigate the effectiveness of the task decomposition. We train an agent without the task decomposition. The results are shown in \tabref{tab:ab:task}. The agent with task decomposition performs better, which illustrates the advantage of the task decomposition.

\subsection{Failure Case}

\figref{fig:failure} shows the failure case. In this case, the foreground object (\ie, car) moves smoothly away from the camera. As the appearance of the foreground object does not change significantly, the VOS algorithm works very well across the whole video sequence. Thus, it is hard for the agent to select the most valuable frame. Our agent achieves comparable performance to the ``Worst'' strategy (\SI{94.43}{\percent} \vs \SI{96.75}{\percent}).

\section{Conclusion}

This paper hypothesizes that the frame with the worst segmentation quality selected in the current interactive VOS is not exactly the best one for annotation. To this end, we formulate the frame recommendation problem as a Markov Decision Process and solve it in the DRL manner. The experimental results on public datasets show that our learned recommendation agent outperforms all baseline strategies without any changes to the underlying VOS algorithms.

\section*{Acknowledgements}

This work was supported by the Special Funds for the Construction of Innovative Provinces in Hunan (2019NK2022), NSFC (61672222, 61932020), National Key R\&D Program of China (2018AAA0100704), Science and Technology Commission of Shanghai Municipality (20ZR1436000), and ``Shuguang Program'' by Shanghai Education Development Foundation and Shanghai Municipal Education Commission. We gratefully acknowledge NVIDIA for GPU donation. We thank the authors of YouTube-VOS for dataset construction.

{
\small
\bibliographystyle{ieee_fullname}
\bibliography{references}
}

\end{document}